\documentclass[runningheads]{llncs}
\usepackage{xcolor}
\usepackage{comment}
\usepackage{graphicx}
\begin{document}
\title{Behavior of k-NN as an Instance-Based Explanation Method}
%
%
\author{Chhavi Yadav \quad
Kamalika Chaudhuri}
\authorrunning{C. Yadav et al.}
%
\institute{UC San Diego \\
\email{cyadav@ucsd.edu} \quad \email{kamalika@cs.ucsd.edu}}
\maketitle              
\begin{abstract}
Adoption of DL models in critical areas has led to an escalating demand for sound explanation methods. Instance-based explanation methods are a popular type that return selective instances from the training set to explain the predictions for a test sample. One way to connect these explanations with prediction is to ask the following counterfactual question - how does the loss and prediction for a test sample change when explanations are removed from the training set? Our paper answers this question for k-NNs which are  natural contenders for an instance-based explanation method. We first demonstrate empirically that the representation space induced by last layer of a neural network is the best to perform k-NN in. Using this layer, we conduct our experiments and compare them to influence functions (IFs) ~\cite{koh2017understanding} which try to answer a similar question. Our evaluations do indicate change in loss and predictions when explanations are removed but we do not find a trend between $k$ and loss or prediction change. We find significant stability in the predictions and loss of MNIST vs. CIFAR-10. Surprisingly, we do not observe much difference in the behavior of k-NNs vs. IFs on this question. We attribute this to training set subsampling for IFs.

\keywords{k-NN  \and Explainability \and Instance-based \and Interpretability \and Representations \and Influence Functions}
\end{abstract}
\section{Introduction}
\label{intro}
Deep Learning (DL) models have shown a great deal of promise in providing high accuracy predictions; however, these accurate predictions come at the cost of high opacity. As they are being adopted in critical applications, it is becoming increasingly important to explain their predictions and interpret them.

A popular class of explanation methods is {\em{instance-based explanations}}. Here the goal is to select instances from the training set to explain why an input is assigned a particular label. k-Nearest Neighbors (k-NN), which provide the top $k$ most similar instances to a test sample, are natural contenders for instance-based explanations. The idea of using k-NNs as an explanation method dates back to the 1990s ~\cite{caruana1999case}. It has recently made a come back like in \cite{rajani2020explaining}. However, the utility of k-NNs as an instance-explanation method has not been investigated thoroughly.

One way to establish the connection between explanations and prediction is to see how the prediction and loss at a test sample change when explanations are removed from the training set. We call this the \textit{`counterfactual question'}. In this work, we take a closer look at k-NNs as instance-based explanations in the context of the above counterfactual question. We compare our k-NN results with removing the most influential training points according to influence function (IF) score from ~\cite{koh2017understanding}. 

To answer this, we first find what is a good representation space in which to choose the nearest neighbors? In the case of a neural network $f$, representation spaces are induced by different layers of $f$. We find that the last layer is the best -- in the sense that k-NN using representations of the last layer approximates the decision boundary of $f$ the best amongst all layers.

Section \ref{exp} discusses our experimental setups and results in detail. Overall, we do see a change in loss and predictions sans the $k$ nearest neighbors.  We observe a significant stability in the predictions and loss of MNIST compared to CIFAR-10. However, we do not observe an increasing trend between the value of $k$ and loss or prediction change. In terms of k-NN vs. IFs, we do not observe much difference in their behaviors. We attribute this to subsampling the training set in the IF case and call for a deeper investigation.

\section{Related Work}
\label{relatedwork}
Instance-based explanations are a popular interpretability tool. Several such methods exist including Counterfactual explanations \cite{wachter2017counterfactual},\cite{karimi2020model}, Prototypes and Criticisms \cite{kim2016examples}, \cite{rdusseeun1987clustering}, k-NNs \cite{caruana1999case}, \cite{rajani2020explaining} and Influence Functions \cite{koh2017understanding}, \cite{koh2019accuracy}, \cite{barshan2020relatif}. Each of them has a different definition and hence a different way of selecting the instance explanations.

Counterfactual explanation method starts with any training instance \& modifies it to reduce the counterfactual loss. The output explanation instance hence need not be from the training set. Prototype based explanation methods first find a representative set of points to describe the data using maximum mean discrepancy and provide the nearest-prototype as explanations for a given test instance. Influence functions estimate the influence of a training sample or set of training samples on a test sample by estimating the change in loss without those training samples. However, they tend to return outliers as noted in \cite{barshan2020relatif}. Given a test sample, k-NNs simply return the k nearest neighbors in a specified representation space as explanations. In this work, we dive deeper into the properties of k-NN as an explanation method.

\section{Experiments \& Results}
\label{exp}
Our goal in this section is to address the following questions:

\begin{itemize}
    \item Which representation space induced by the different layers of a neural network $f$ is best for finding k-NNs?
    \item How does removing the top $k$ nearest neighbors (offered as explanations) from the training data affect the prediction on a test input?
\end{itemize}

These questions are considered in the context of two standard datasets -- MNIST and CIFAR-10. For MNIST we use a custom model as shown in table \ref{tb1:Percent}, while for CIFAR-10 we used a Wide-ResNet ~\cite{zagoruyko2016wide} with depth and width 10.

\subsection{The Choice of Representation}

Since k-NN is a distance based method, the distance function and representation space in which these distances are found, are two significant aspects that affect its output.
In a neural network $f$, each layer induces its own representation space $\phi_{layer}$, giving k-NN an option to choose amongst these.


\paragraph{\textbf{Label Agreement}} : We say that label agreement happens for a test point when the label predicted by nearest neighbor and the neural network are the same. We find the fraction of test points for which label agreement happens in different layers of $f$. As this fraction approaches 1, gap between the decision boundaries of k-NN and $f$ reduces, becoming approximately the same, when tested with large number of samples. If this occurs, then k-NN is a global approximator of the neural network.


\paragraph{\textbf{Setup}} : We take 1000 samples from the test set of each dataset, with equal contributions from all classes. We find the 1-NNs from the training set using representations from various layers of the neural network. We compare labels from 1-NN to those predicted by the neural network. We use cosine distance function.

\paragraph{\textbf{Result}} : From tables \ref{tb1:Percent} and \ref{tb2:Absolute_cifar} it is clear that as we go from first layer to the last, the fraction of points with label agreement increases. For MNIST, the fraction approaches 1 at the last layer, while for CIFAR-10, it approaches 0.9. Since k-NN in the last layer approximates decisions of $f$ the best, we use $\phi_{last}$ as our k-NN representation space.

\begin{table}[ht]
  \centering
  \caption{\label{tb1:Percent} Fraction of points for which the label predicted by 1-NN and neural network are same on MNIST}
  \smallskip
    \begin{tabular}{lr}
    \hline
     Layer         &   Fraction\\
    \hline
     conv\_relu         &  0.962 \\
     conv\_relu\_maxpool         & 0.977 \\
     conv\_relu    &  0.979 \\
     conv\_relu\_maxpool         &  0.983 \\
     fc           &  0.995 \\
     relu\_drop\_fc           &  0.997  \\
     relu\_fc           &  0.998 \\
    \hline
    \end{tabular}
\end{table}

\begin{table}[ht]
  \centering
  \caption{\label{tb2:Absolute_cifar} Fraction of points for which the label predicted by 1-NN and neural network are same on CIFAR-10}
  \smallskip
  \begin{tabular}{lr}
    \hline
     Layer         &  Fraction\\
    \hline
     conv          &    0.328\\
     block1\_block2\_block3        &  0.756\\
     batchnorm\_relu\_avgpool          &   0.877\\
     fc            &     0.881\\
    \hline
    \end{tabular}
\end{table}

\subsection{Effect of Removing Explanatory Instances}
Given a test sample $x$, our k-NN explainer (in $\phi_{last}$) would return the top $k$ nearest neighbors of $x$ from the training set as explanations. In this section, we want to understand the effect of removing these $k$ samples from the training set on the prediction and loss at $x$.
In ~\cite{koh2017understanding}, the authors define highly influence points to be those training points that when removed from the training set, lead to the largest change in loss at $x$. Therefore, we compare our results to influence functions.

\paragraph{\textbf{Setup}} : We take 500 random samples from the test set and find their top $k$ nearest neighbors from the training set using $\phi_{last}$ and cosine distance function. $k \in [1,5,10,15,20]$. For each test sample $x$, we remove these k nearest neighbors from the training set and retrain a model from scratch using the new training set, keeping other parameters fixed. We note the loss change (LC) and predicted labels for the test samples before and after retraining. A Positive LC implies that the loss at $x$ increased due to removal of the $k$ nearest neighbors from the training set. Similarly, a change in predicted label before and after retraining, which we call a label flip, implies some effect was caused by removal of the k nearest neighbors.

We compare our results to Influence Functions (IFs). For each test sample $x$, we find the top $k$ influential points using the definition in ~\cite{koh2017understanding} and follow the same procedure as above. Calculating influence is an expensive task since for each test sample, it takes $O(np + rtp$) time, where $rt = O(n)$, $n$ = size of dataset, $r$ = repeats, $t$ = iterations \& $p$ = model parameters, as mentioned in ~\cite{koh2017understanding}. Hence, due to computational constraints, we use some commonplace tricks. We test using 100 random test samples. When searching for influential samples in the training set we look in a subsampled training dataset of size 1000. However, we retrain with the complete training set minus the top $k$ influential samples. We freeze all but last layer weights.

\paragraph{\textbf{Results}} :
As is evident from Table \ref{knn_mnist_lc} and \ref{lf}, we do see a loss and prediction change when top $k$ nearest neighbors are removed from training set. However, there is no particular increasing trend between the value of $k$ and loss change or label flips. This is counter-intuitive since one would expect removing more samples would lead to potentially worse learning and hence higher loss change. A plausible explanation can be that there is enough information in the rest of the training samples to learn. Hence, removing a few does not matter much, unless the test sample is noisy or an outlier.

We also note that predictions on MNIST are more stable than CIFAR-10, in the sense that the loss change and percentage of label flips from table \ref{lf} for MNIST are way lower than that for CIFAR-10. This reiterates the belief that MNIST is an easier dataset and has less intraclass diversity in the samples. 

Contrasting the results of k-NN and Influence functions, we do not observe much difference in behavior. Removing top $k$ nearest neighbors vs. influential points has almost similar effects. This maybe due to subsampling the training set for influence functions for computational reasons \& calls for further investigation.

\begin{table}
\centering
\caption{k-NN : For various k, loss change (LC) at test samples between retrained and original models. Average(Avg.) LC is the average across all test samples, Avg. +ve \& -ve LC are the average positive \& negative loss change across respective test samples, Max. \& Min. LC note the maximum \& minimum loss change across all test samples, +ve LC \% notes the percentage of test samples with a $\geq$ 0 LC.  }
\label{knn_mnist_lc}
\smallskip
\begin{tabular}{|l|l|c|c|c|c|c|c|}
    \hline
    Dataset & k &    Avg. LC  &   Avg. +ve LC &   Avg. -ve LC & Max. LC & Min. LC &  +ve LC \% \\
    \hline

        & 1 & 0.009   $\pm$    0.177 &     0.015 $\pm$    0.180 &    -0.040 $\pm$    0.146 & 2.622  & -0.752 & 89.6 \\
        & 5 & 0.019   $\pm$    0.499 &     0.026 $\pm$    0.525 &    -0.040 $\pm$    0.148 & 11.107 & -0.786 & 89.2 \\
    MNIST & 10 & 0.029  $\pm$    0.452 &     0.036 $\pm$    0.473 &    -0.036 $\pm$    0.133 & 9.392  & -0.624 & 90.4 \\
        & 15 & 0.024  $\pm$    0.327 &     0.031 $\pm$    0.340 &    -0.040 $\pm$    0.136 & 5.375  & -0.553 & 91.0\\
        & 20 & 0.034  $\pm$    0.535 &     0.043 $\pm$    0.560 &    -0.050 $\pm$    0.163 & 8.354  & -0.747 & 90.4 \\\hline
        & 1  &    -0.407 $\pm$       2.018 &         0.751 $\pm$       1.597 &        -0.908 $\pm$       1.975 &     8.514 &   -19.012 &     30.2 \\
        & 5  &    -0.446 $\pm$       1.866 &         0.647 $\pm$       1.238 &        -0.928 $\pm$       1.893 &     6.169 &   -15.23  &     30.6\\
    CIFAR-10   & 10 &    -0.342 $\pm$       2.643 &         1.009 $\pm$       3.227 &        -0.954 $\pm$       2.057 &    35.885 &   -19.441 &     31.2\\
        & 15 &    -0.075 $\pm$       3.134 &         1.516 $\pm$       4.383 &        -0.851 $\pm$       1.844 &    44.61  &   -19.101 &     32.8  \\
        & 20 &    -0.526 $\pm$       1.954 &         0.483 $\pm$       1.084 &        -0.965 $\pm$       2.082 &     6.984 &   -18.344 &     30.3\\
    \hline
\end{tabular}
\end{table}

\begin{table}
\centering
\caption{Influence Functions : For various k, loss change (LC) at test samples between retrained and original models. Average(Avg.) LC is the average across all test samples, Avg. +ve \& -ve LC are the average positive \& negative loss change across respective test samples, Max. \& Min. LC note the maximum \& minimum loss change across all test samples, +ve LC \% notes the percentage of test samples with a $\geq$ 0 LC.  }
\label{inf_mnist_lc}
\smallskip
\begin{tabular}{|l|l|c|c|c|c|c|c|}
    \hline
    Dataset & k &    Avg. LC  &   Avg. +ve LC &   Avg. -ve LC & Max. LC & Min. LC &  +ve LC \% \\
    \hline

        & 1 &  -0.00019  $\pm$      0.002 &       1.3e-05   $\pm$       8e-05     &      -0.00285 $\pm$       0.007 &    0.0006 &    -0.02  &         93 \\
        & 5 &   7e-05   $\pm$       0.003 &       0.00029 $\pm$       0.002 &      -0.00346 $\pm$       0.008 &    0.0185 &    -0.02  &         94 \\
    MNIST & 10 &  -0.00021 $\pm$       0.002 &       2e-06       $\pm$       2e-05     &      -0.00232 $\pm$       0.006 &    0.0001 &    -0.02  &         91 \\
        & 15 &  -0.00018 $\pm$       0.002 &       4e-06       $\pm$       3e-05     &      -0.00229 $\pm$       0.006 &    0.0003 &    -0.018 &         92 \\
        & 20 &  -0.00018 $\pm$       0.002 &       2e-05   $\pm$       1e-04     &      -0.00343 $\pm$       0.008 &    0.0007 &    -0.02  &         94 \\\hline
        & 1 &    -0.43  $\pm$       1.544 &         0.464 $\pm$       1.039 &        -0.87  $\pm$       1.562 &     5.373 &    -8.501 &         33 \\
        & 5 &    -0.349 $\pm$       1.334 &         0.666 $\pm$       1.187 &        -0.688 $\pm$       1.201 &     4.416 &    -4.998 &         25 \\
        CIFAR-10 & 10 &    -0.447 $\pm$       1.637 &         0.494 $\pm$       1.037 &        -0.89  $\pm$       1.68  &     4.319 &    -8.891 &         32 \\
        & 15 &    -0.539 $\pm$       1.623 &         0.378 $\pm$       0.948 &        -1.077 $\pm$       1.694 &     4.323 &    -6.95  &         37 \\
        & 20 &    -0.475 $\pm$       1.351 &         0.314 $\pm$       0.699 &        -0.797 $\pm$       1.418 &     3.398 &    -7.776 &         29 \\
    \hline
\end{tabular}
\end{table}

\begin{table}[htp]
\centering
\caption{Percentage of label flips over test samples, between the retrained and original models when the top $k$ nearest neighbors \& the top $k$ influential samples are removed from the training set.}
\label{lf}
\smallskip
\begin{tabular}{|l|c|c|c|c|}
\hline
k & MNIST k-NN & MNIST IF & CIFAR k-NN & CIFAR IF \\
\hline
1 &             0.8 &  0 & 21.2 & 23\\
5 &             0.6 &  0 & 23.2 & 19\\
10 &             1 &  0 & 22.8 & 22\\
15 &             0.6  & 0 & 23.4 & 19\\
20 &             0.8 & 0 & 19.48 & 21\\
\hline
\end{tabular}
\end{table}

\section{Conclusion}
\label{disnconc}
We present an investigation into the behavior of k-NN as an instance-based explanation method. First, we demonstrate that representation space induced by the last layer of a neural network is the best to find neighbors in. Following this, we try to connect explanations and predictions by probing the counterfactual question - what happens when the k nearest neighbors of a test sample are removed from the training set? How are the loss and prediction affected?

From our experiments we do see a change in loss and predictions. However, we do not find signs of an increasing relationship between the value of $k$ and loss or prediction change. We also find evidence for stable predictions on MNIST than CIFAR-10.

We do not observe much difference between the behavior of k-NN vs. IFs. We believe that this could be due to subsampling the training set for computational constraints. A future direction can be to investigate this more thoroughly with the complete training set, though it is a difficult task in reality.

We conclude by saying that when choosing an explanation method, it's of utmost importance to think carefully about what the application is and if the properties of the explanation method match the requirements. In line of this, we call for a deeper investigation into the properties of explanation methods.

\bibliographystyle{splncs04}
\bibliography{ref}

\end{document}